\pdfoutput=1
\documentclass[11pt]{article}
\usepackage[final]{EMNLP2023}
\usepackage{graphicx}
\usepackage{times}
\usepackage{latexsym}
\usepackage{multirow}
\usepackage[T1]{fontenc}
\usepackage[utf8]{inputenc}
\usepackage{microtype}
\usepackage{inconsolata}

\title{RSM-NLP at BLP-2023 Task 2: Bangla Sentiment Analysis\\ using Weighted and Majority Voted Fine-Tuned Transformers}
\author{Pratinav Seth\thanks{\hspace{0.25cm}Dept. of Data Science \& Computer Applications}\hspace{0.5cm}Rashi Goel\thanks{\hspace{0.25cm}Dept. of Computer Science \& Engineering }\hspace{0.5cm}Komal Mathur \footnotemark[2] \hspace{0.1cm} \thanks{\hspace{0.25cm}Authors have contributed equally to this work}\hspace{0.5cm}Swetha Vemulapalli\thanks{\hspace{0.25cm}Dept. of  Information and Communication Technology} \hspace{0.1cm} \footnotemark[3] \\
         Manipal Institute of Technology\\ Manipal Academy of Higher Education, Manipal, India\\
 \texttt{\{seth.pratinav,rashigoel2017,komlixmathur,swetha.vemulapalli.3\}@gmail.com
 }
 }

\begin{document}
\maketitle
%\footnotetext[1]{Both authors contributed equally.}
\begin{abstract}
This paper describes our approach to submissions made at Shared Task 2 at BLP Workshop - Sentiment Analysis of Bangla Social Media Posts\cite{blp2023-overview-task2, islam-etal-2021-sentnob-dataset, hasan2023zero}.
Sentiment Analysis is an action research area in the digital age. With the rapid and constant growth of online social media sites and services and the increasing amount of textual data, the application of automatic Sentiment Analysis is on the rise. However, most of the research in this domain is based on the English language. Despite being the world's sixth most widely spoken language, little work has been done in Bangla. 
This task aims to promote work on Bangla Sentiment Analysis while identifying the polarity of social media content by determining whether the sentiment expressed in the text is Positive, Negative, or Neutral.
Our approach consists of experimenting and finetuning various multilingual and pre-trained BERT-based models on our downstream tasks and using a Majority Voting and Weighted ensemble model that outperforms individual baseline model scores. Our system scored 0.711 for the multiclass classification task and scored 10th place among the participants on the leaderboard for the shared task. Our code is available at \url{https://github.com/ptnv-s/RSM-NLP-BLP-Task2}

\end{abstract}

\section{Introduction}
In the era of a high influx of social media platforms, blogs, and online reviews, sentiment analysis has become the need of the hour. Also known as opinion mining, sentiment analysis is a computational linguistic task that is aimed at determining whether a text contains a positive, negative, or neutral sentiment behind it \cite{khan2020sentiment}
Sentiment analysis has diverse uses, including preventing adolescent suicide by detecting cyberbullying and mitigating unjust actions that target specific communities through hate speech detection, among numerous other applications \cite{islam2020sentiment}.
Approximately 284.3 million people worldwide speak Bangla as their primary language. Individuals speaking Bangla increasingly engage in social media platforms like Instagram, Facebook, Reddit, and Twitter and express opinions on micro-blogging platforms, commenting on news portals and online shopping. However, analyzing vast volumes of rapidly generated data in the digital age is a very tedious job to do. This is where sentiment analysis can be applied \cite{hassan2016sentiment}.
Most sentiment analysis research predominantly focuses on English, leaving Bangla Sentiment analysis in its nascent stages. Recently, some works have addressed this issue. However, none of these studies have fully embraced the different perspectives of Bangla.
%The research needs to take into consideration the standardization of Bangla, Banglish (mixing Bangla with English), and Romanized Bangla.\cite{hassan2016sentiment}
\begin{table}[pt]
\begin{tabular}{cc}
\hline
Text               & Label                     \\ \hline
\multirow{2}{*}{\includegraphics[width=0.7\linewidth]{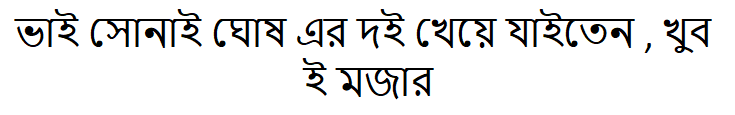}} & \multirow{2}{*}{Positive} \\
                   &                           \\
\multirow{2}{*}{\includegraphics[width=0.7\linewidth]{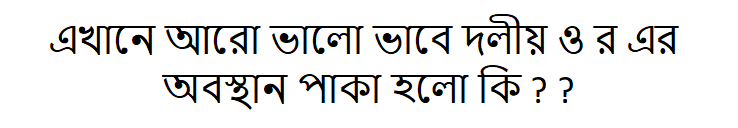}} & \multirow{2}{*}{Neutral}  \\
                   &                           \\
\multirow{2}{*}{\includegraphics[width=0.7\linewidth]{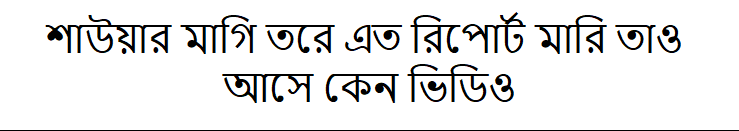}} & \multirow{2}{*}{Negative} \\
                   &        \\ \hline                          
\end{tabular}
\caption{ Text Samples from the Training dataset, with labels as either Positive, Neutral or Negative}\label{tab1}
\end{table}

To address this problem, we present our contributions to Shared Task 2 at BLP Workshop - Sentiment Analysis of Bangla Social Media Posts. This task aims to detect the polarity associated with a given social media text. This multiclass classification task involves determining whether the sentiment expressed in the text is Positive, Negative, or Neutral. For this problem statement, we have conducted various experiments using multi-lingual berts \cite{bhattacharjee-etal-2022-banglabert,Sanh2019DistilBERTAD,das2022data,Sagor_2020} and various pre-trained transformers \cite{DBLP:journals/corr/abs-1907-11692} by fine-tuning them on downstream tasks. We also apply Majority Voting and Weighted ensembling on the top-k models to show how these methods affect the models' performance and how an ensemble of these models performs better than the individual baselines.
\section{Background}

\subsection{Problem and Data Description}
The EMNLP 2023 Bangla Workshop Task 2: Sentiment Analysis of Bangla Social Media Posts \cite{blp2023-overview-task2, islam-etal-2021-sentnob-dataset, hasan2023zero} aims to detect the polarity of the sentiment associated with a given text extracted from social media. 
From the entire set of labels, over 14,000 were classified as negative, approximately 12,000 as positive, and roughly 6,000 as neutral, as indicated in the distribution chart in Figure \ref{fig:example} and a few samples of the Dataset are shown in Table~\ref{tab1}.
The dataset includes the MUltiplatform BAngla SEntiment (MUBASE) dataset and the SentNob dataset \cite{islam-etal-2021-sentnob-dataset}. SentNob comprises public comments from social media on news and videos across 13 domains, such as agriculture, politics, and education. It is manually annotated with a moderate agreement score of 0.53. On the other hand, MUBASE is a sizable compilation of multi-platform data, including Facebook posts and tweets, each manually tagged for sentiment polarity. These datasets provide a comprehensive and diverse landscape for studying Bangla sentiment analysis.
\subsection{Previous Works}
\subsubsection{Sentiment Analysis}
Sentiment analysis is an NLP task that uses computational methods to determine and extract the emotional tone expressed in a piece of text \cite{HOGENBOOM201443}.
There are several different approaches to sentiment analysis. Early sentiment analysis approaches primarily employed rule-based methods and lexicon-based techniques \cite{obaidat2015enhancing} to determine the sentiment context of texts.
One of the significant areas of application of Sentiment Analysis is in Social Media Posts as in \cite{tang2014building} and \cite{taboada2011lexicon}, a sentiment lexicon with a linguistic rule-based approach was used to create a sentiment detection mechanism from tweets\cite{reckman2013teragram}.
Following this, contemporary advancements have introduced machine learning and deep learning techniques that significantly boost accuracy by extracting intricate patterns from annotated datasets. 
Due to human language's complexity and sentiment expression nuances, it is a challenging task. The accuracy of the task may be improved by using larger datasets, more complex and 
fine-tuned models \cite{hassan2016sentiment}, ensembling, etc.
Modern approaches leverage large-scale Pre-trained Language Models (PLMs), such as Transformers, BERTs \cite{devlin2018bert}, 
and NLUs \cite{bender2020climbing}, alongside refined fine-tuning mechanisms\cite{hasan2023zero}. They excel at capturing the intricate associations between words within the text and their corresponding polarity.
In today's world, with the introduction of free-to-use models like ChatGPT, sentiment analysis has opened to new possibilities \cite{Wang2023IsCA}.
\subsubsection{Bangla Language Processing}
\begin{figure}[pt]
  \centering
\includegraphics[width=1\linewidth]{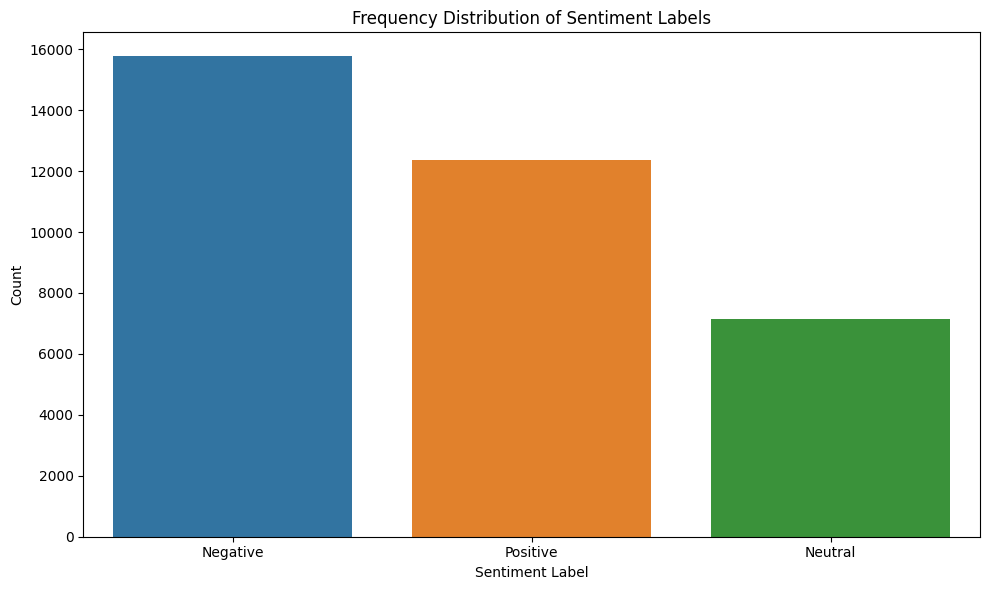} % Adjust the width as needed
  \caption{Frequency of Task 2 labels in training set}
  \label{fig:example}
\end{figure}
The Bangla language is the 7th most spoken language, with 265 million speakers worldwide \cite{9751052}. However, since English is the predominant language used for technical knowledge, journals, and documentation, many Bangla-speaking people face hurdles in utilizing these resources.
Research on Bangla Natural Language Processing (BNLP) began in the early 1990s, focusing on rule-based lexical and morphological analysis \cite{alam2021review}. 
From the modeling perspective, most earlier endeavors are either rule-based, statistical, or classical machine learning-based approaches\cite{kudo2001chunking}. As for the sequence tagging tasks, such as NER and G2P, the algorithms, including
Hidden Markov Models (HMMs) \cite{brants2000tnt}, Conditional Random Fields (CRFs) \cite{lafferty2001conditional}, Maximum Entropy
(ME) \cite{ratnaparkhi1996maximum} and Maximum Entropy Markov Models (MEMMs) \cite{mccallum2000maximum} have been used successfully.
It is only very recently that a small number of studies have explored deep learning-based approaches.
As depicted in \cite{alam2021review}, there has been significant work in resource and model development in Bangla sentiment analysis. In \cite{das2010sentiwordnet}, the authors proposed a computational technique of generating an equivalent SentiWordNet (Bangla) from publicly available English sentiment lexicons and an English-Bangla bilingual dictionary with few easily adaptable noise reduction techniques. However, with the Introduction of BERTs many works focused on fine-tuning multilingual BERTs \cite{9044317,das-etal-2021-emotion}, but BanglaBERT \cite{Sagor_2020} being the first model pre-trained on Bangla text corpus.
\subsubsection{Bangla Sentiment Analysis}
Sentiment analysis is a tool to extract the emotional tone of the text. It is used for cyberbullying detection, hate speech mitigation and market research. Bangla is the 7th most spoken language, and sentiment analysis for Bangla is still in its early stages. The first attempt to perform sentiment analysis in the context of Indian Languages, including Bangla, was done as recently as in 2015 \cite{patra2015shared}. 
\\The lack of accurately annotated data is one of the biggest bottlenecks to advancing Bangla Sentiment Analysis. \cite{islam-etal-2021-sentnob-dataset} and \cite{rahman2018datasets} describe the creation of datasets for this purpose. 
A word2vec model was tuned with word co-occurrence scores for sentiment analysis in \cite{al2017sentiment}, achieving an accuracy of 75.5\%. In \cite{wahid2019cricket}, aspect-based sentiment analysis data was examined, boasting a remarkable 95\% accuracy. 
However, challenges were encountered when rephrasing common and proper nouns in Bangla.
Among most studies, however, transformer models have consistently outperformed other algorithms and models, inciting a significant amount of research into the area. 
In \cite{chowdhury2019analyzing}, Opinion Mining was conducted on a dataset of 4,000 manually translated Bangla movie reviews, with the objective of classifying them as positive or negative. The LSTM approach had achieved an accuracy of 82.42\%. A Bi-LSTM architecture was applied by \cite{sharfuddin2018deep} to a labeled dataset of 10,000 Facebook comments in Bangla, resulting in an accuracy of 85.67\%. However, the study faced significant data preprocessing difficulties.
In \cite{tripto2018detecting}, a combination of CNN and LSTM was employed to extract six distinct emotions from various types of Bangla YouTube video comments. The reported accuracies were 65.97\% and 54.24\% for three and five-label sentiment classification, respectively. A common issue faced by authors while using CNNs was that proper tuning between layers could not be achieved.
In another study \cite{hossain2020sentiment}, 1000 online restaurant reviews were collected from the Foodpanda website for performing SA and deployed, thus combining CNN with LSTM architecture with a 300 dimensional Word2Vec pretrained model having validation accuracy of 75.01\%. \cite{rezaul2020classification} developed a novel word embedding system for Bangla texts, BanglaFastText, incorporating it into a Multichannel Convolutional LSTM (MConv-LSTM). 
In \cite{islam2020sentiment} authors performed SA on 1002 public comments from newspapers with the help of the BERT pretrained model and achieved accuracy on GRU at 71\% on 2 class sentiments. 
In \cite{iccit2020Arid}, the performance of multiple classical machine learning algorithms and deep learning models were compared on several sentiment-labeled datasets, showing that pre-trained transformer models such as BERT and XLM-RoBERTa yielded the highest scores.

\section{System Overview}
We conducted extensive experiments for the given task involving Bangla Sentiment analysis. We fine-tuned various multilingual and pre-trained transformer architectures, including BERT \cite{kenton2019bert}, DistillBERT \cite{sanh2019distilbert}, RoBERTa \cite{liu2019roberta}, and Various Pre-Trained BERT models \cite{das2022data,Sagor_2020} on our downstream task of polarity classification.
We shortlist the top-k model based on the performance metrics and ensemble the predictions using Majority Voted and Weighted Ensemble.
\subsection{Fine-Tuning Transformers}
\begin{table}[pt]
\begin{tabular}{ccccc}
\hline
\textbf{Model} & \textbf{Acc.} & \textbf{Pre.} & \textbf{Rec.} & \textbf{F1} \\ \hline
\begin{tabular}[c]{@{}c@{}}RoBERTa\\ (Base)\end{tabular} & 0.550 & 0.544 & 0.550 & 0.550 \\
\begin{tabular}[c]{@{}c@{}}Distill\\ BERT\end{tabular} & \textbf{0.701} & \textbf{0.687} & \textbf{0.701} & \textbf{0.701} \\
\begin{tabular}[c]{@{}c@{}}HF-PT\\ BERT-1\end{tabular} & 0.672 & 0.679 & 0.672 & 0.672 \\
\begin{tabular}[c]{@{}c@{}}HF-PT\\ BERT-2\end{tabular} & 0.639 & 0.630 & 0.639 & 0.639 \\
\begin{tabular}[c]{@{}c@{}}HF-PT\\ BERT-3\end{tabular} & 0.669 & 0.671 & 0.669 & 0.669 \\
\begin{tabular}[c]{@{}c@{}}Bangla\\ BERT\\ (Small)\end{tabular} & 0.657 & 0.649 & 0.657 & 0.657 \\
\begin{tabular}[c]{@{}c@{}}Bangla\\ BERT\\ (Large)\end{tabular} & 0.693 & 0.684 & 0.693 & 0.693 \\
\begin{tabular}[c]{@{}c@{}}Bangla\\ BERT\\ (Base)\end{tabular} & \textbf{0.701} & \textbf{0.687} & \textbf{0.701} & \textbf{0.701} \\
\begin{tabular}[c]{@{}c@{}}Banglish\\ BERT\end{tabular} & 0.684 & 0.672 & 0.684 & 0.684 \\ \hline
\end{tabular}
\caption{Results of Base-Models on Test-Set of Shared-Task Dataset where Acc. is Accuracy, Pre. is Precision, Rec. is Recall \& F1 refers to F1-Score}\label{tab2}
\end{table}
We used multiple transformer architectures to observe the effect of the model architecture and the pre-trained dataset on the downstream task.
For multiclass classification, we added a linear layer acting as a classification head to fine-tune the models for the multiclass classification. 

We have used various models for our experiments, including 
\textbf{BERT} \cite{kenton2019bert}, a transformer-based language model that creates representations of text by combining both left and right contexts with Masked Language Modeling and Next Sentence Prediction being pre-training tasks.
\textbf{RoBERTa} \cite{liu2019roberta} is a faster variation of BERT. 
\textbf{DistilBERT (multilingual cased)} \cite{sanh2019distilbert} is a distilled version of the multilingual Bert with pretraining on Wikipedia data in 104 languages. 
\textbf{BanglaBERT} \cite{Sagor_2020} referred to as HF-PT-BERT-2 in Table \hyperref[table1]{1} is a pretrained BERT trained on the Bangla common crawl dataset and the Bangla Wikipedia Dump Dataset.
\textbf{Indic-abusive-allInOne-MuRIL} \cite{das2022data} is a model finetuned from the MuRIL \cite{khanuja2021muril} and multilingual BERT models, trained to detect abusive speech using multiple datasets in 8 Indian languages.
\textbf{Bengali-abusive-MuRIL} \cite{das2022data} is also finetuned from MuRIL \cite{khanuja2021muril}, trained specifically on the Bangla abusive speech dataset. These have been referred to as HF-PT-BERT-1 and HF-PT-BERT-3 in Table \hyperref[table1]{1}, respectively. 
\textbf{BanglaBERT} \cite{bhattacharjee-etal-2022-banglabert}is a fine-tuned ELECTRA \cite{clark2020electra} model which is trained on Bangla Wikipedia dump dataset as well as data from 110 Bangla websites.
\textbf{BanglishBERT}\cite{bhattacharjee-etal-2022-banglabert}is similar to BanglaBERT; instead, it was trained on both English and Bangla data to allow zero-shot cross-lingual transfer.
\subsection{Ensembling Predictions}

To increase the overall performance of the predictions and robustness of the predictive model, models were first individually tuned on the downstream task dataset. The predictions from these models were combined using the two ensembling methods on top-3,top-5, and all model predictions:\\
\textbf{Majority Voting}: The most frequently occurring prediction from all the models for each training instance was chosen as the final label.\\
\textbf{Weighted}: Each model was assigned a weight based on its accuracy score on the training dataset. Each model voted on the prediction class with its weight, and the prediction with the highest final vote was chosen as the final label.
\begin{equation}
\label{sec:wtens}
    y_{i} = argmax(\sum_{j=1}^k a_{j}.p_{ij})
\end{equation}
Here, $y_{i}$ denotes the Weighted ensemble prediction of the ith sample, $p_{ij}$  the ith probabilistic prediction for each polarity made by the jth model, $a_{j}$  the accuracy of the jth model on the training set and k is the number of models being considered for the ensemble.

\section{Experiments \& Results}
The dataset used for the task is organized in 3 columns, with id, text, and label. It has also been partitioned into a train set with 35266 samples, a dev set with 3935 samples, and a dev-test set with 3427 samples. The distribution in the training set is shown in Figure \ref{fig:example}.

The preprocessing pipeline before model training included padding, tokenizing, and truncating text data to ensure uniformity and manage lengthy inputs.
We used the AdamW optimizer, a learning rate of $2$x$10^{-5}$ and a batch size of 32 over 32 epochs was chosen to strike a balance between convergence speed and stability with a maximum sequence length of 512 tokens used with Huggingface AutoTokenizer to tokenize the data.

We evaluated models using four metrics: accuracy, precision, recall, and F1-score. F1-score is a good metric for imbalanced datasets because it takes into account both precision and recall.

The results of our experiments over the official Test set are shown in Table~\ref{tab2} \&~\ref{tab3}. For Individual Models as shown in Table~\ref{tab2} we observe DistilBERT and BanglaBERT(Base) show the best performance on the test data, with an F1-Score of 0.701.

We did an ensemble of both types (Majority-Voted and Weighted) with the top 3 ( BanglaBERT \cite{Sagor_2020}, BanglishBERT, HF-PT-BERT-1 \cite{das2022data} ), top 5 (HF-PT-BERT-2, BanglishBERT, HF-PT-BERT-1 
\cite{das2022data} , BanglaBERT(Base), HF-PT-BERT-3 \cite{das2022data} ) and lastly using all the models. 
As in Table~\ref{tab3} for ensembles, we observe that the majority ensemble shows a better performance in general as compared to the weighted models. The majority voted ensemble using predictions from all the models had the highest F1 score of 0.711. Furthermore, an ensemble of 3 models yielded almost optimal results. The use of more than three models resulted in a marginal increase in performance but significantly increased resource utilization. Thus, the use of more than three models seems unproductive. 
\section{Conclusion}
\begin{table}[pt]
\begin{tabular}{cccccc}
\hline
\textbf{Method} & \textbf{Top} & \textbf{Acc.} & \textbf{Prec.} & \textbf{Rec.} & \textbf{F1} \\ \hline
\multirow{3}{*}{\begin{tabular}[c]{@{}c@{}}Majo\\ -rity \\ Voted\end{tabular}} & 3 & 0.706 & 0.692 & 0.706 & 0.706 \\
 & 5 & 0.707 & 0.694 & 0.707 & 0.707 \\
 & All & \textbf{0.711} & \textbf{0.695} & \textbf{0.711} & \textbf{0.711} \\
\multirow{3}{*}{\begin{tabular}[c]{@{}c@{}}Weig\\ -hted\end{tabular}} & 3 & 0.703 & 0.691 & 0.703 & 0.703 \\
 & 5 & 0.703 & 0.692 & 0.703 & 0.703 \\
 & All & 0.708 & \textbf{0.695} & 0.708 & 0.708 \\ \hline
\end{tabular}
\caption{Results of ensemble models on Test-Set of Shared-Task Dataset where Method is the method of ensembling, Top refers to top-k models chosen, Acc. is Accuracy, Pre. is Precision, Rec. is Recall \& F1 refers to F1-Score}\label{tab3}
\end{table}
In this work, we benchmarked various multilingual and pre-trained BERT-based models - RoBERTa\cite{DBLP:journals/corr/abs-1907-11692}, DistillBERT\cite{Sanh2019DistilBERTAD}, BanglaBERT\cite{bhattacharjee-etal-2022-banglabert}, BanglishBERT\cite{hasan-etal-2020-low} and Various Pre-Trained BERT models \cite{das2022data,Sagor_2020} for Bangla Sentiment Analysis \cite{blp2023-overview-task2, islam-etal-2021-sentnob-dataset, hasan2023zero} while identifying the polarity of social media content by determining whether the sentiment expressed in the text is Positive, Negative, or Neutral as our downstream tasks and using a Majority Voting and Weighted ensemble model that outperforms individual baseline model scores. 

Our system achieved a micro F1-Score of 0.711 for the multiclass classification task and scored 10th among the participants on the leaderboard for the shared task.
\section{Acknowledgments}
We would like to thank the Research Society MIT, an undergraduate interdisciplinary technical society of Manipal Institute of Technology, Manipal Academy of Higher Education, Manipal, India, for supporting our research.
\bibliography{custom}
\bibliographystyle{acl_natbib}
\appendix
%\section{Appendix}
%\label{sec:appendix}
\end{document}